\pdfoutput=1
\def\year{2021}\relax
\documentclass[letterpaper]{article} 
\usepackage{aaai21}  
\usepackage[hyphens]{url}  
\usepackage{graphicx} 
\usepackage[numbers]{natbib}  
\usepackage{caption} 
\usepackage[dvipsnames]{xcolor}

\usepackage{subcaption}
\usepackage[ruled,vlined]{algorithm2e}
\usepackage{bm}
\graphicspath{{./figs/}{./figs/cylinder-results/}{./figs/bracket-results/}{./figs/chair-results/}}
\title{Surrogate Modeling for Physical Systems \\ with Preserved Properties and Adjustable Tradeoffs}
\author{Randi Wang\textsuperscript{\rm 1}, Morad Behandish\textsuperscript{\rm 2}\thanks{Corresponding Author}
    \\
}
\affiliations{
    \textsuperscript{\rm 1}University of Wisconsin-Madison,
    \textsuperscript{\rm 2}Palo Alto Research Center (PARC)\\

    \textsuperscript{\rm 1}randi.wang@wisc.edu,
    \textsuperscript{\rm 2}moradbeh@parc.com~~

}

\begin{document}

\maketitle

\begin{abstract}
Determining the proper level of details to develop and solve physical models is usually difficult when one encounters new engineering problems. Such difficulty comes from how to balance the time (simulation cost) and accuracy for the physical model simulation afterwards. We propose a framework for automatic development of a family of surrogate models of physical systems that provide flexible cost-accuracy tradeoffs to assist making such determinations. We present both a model-based and a data-driven strategy to generate surrogate models. The former starts from a high-fidelity model generated from first principles and applies a bottom-up model order reduction (MOR) that preserves stability and convergence while providing a priori error bounds, although the resulting reduced-order model may lose its interpretability. The latter generates interpretable surrogate models by fitting artificial constitutive relations to a presupposed topological structure using experimental or simulation data. For the latter, we use Tonti diagrams to systematically produce differential equations from the assumed topological structure using algebraic topological semantics that are common to various lumped-parameter models (LPM). The parameter for the constitutive relations are estimated using standard system identification algorithms. Our framework is compatible with various spatial discretization schemes for distributed parameter models (DPM), and can supports solving engineering problems in different domains of physics.

\end{abstract}

\section{Introduction}
\begin{figure} 
	\centering\includegraphics[width=0.8\linewidth]{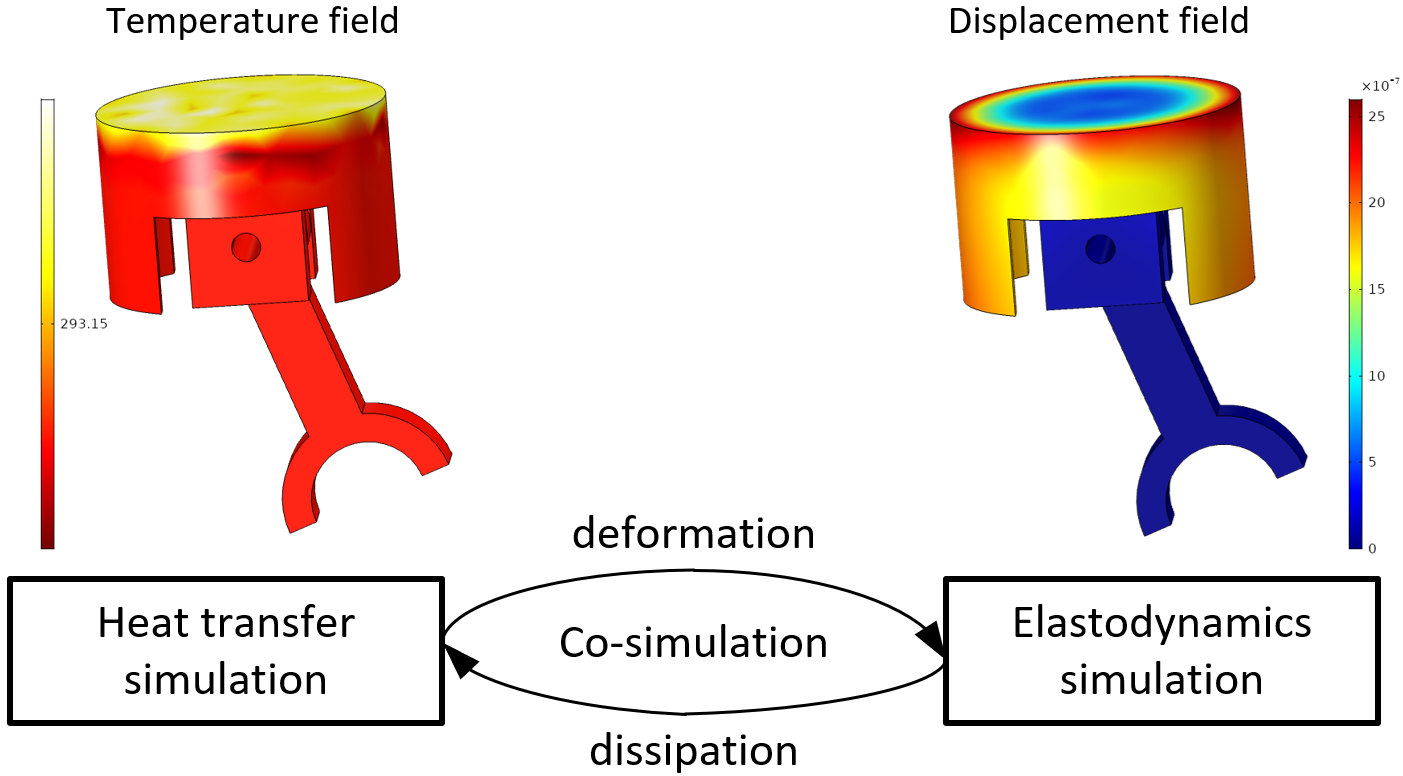}
	\caption{Displacement and thermal fields of a piston simulated by solving its finite element model}
	\label{fig:piston_comsol}
\end{figure}

Determining the appropriate level of granularity (LOG) to model a physical system for computer-aided simulation is a nontrivial problem, as cost-accuracy tradeoffs are rarely clear upfront. Generally, when encountered with a new practical problem, physicists and engineers attempt to model the system as accurately as possible, starting from the first principles. For example, Fig. \ref{fig:piston_comsol} illustrates a multiphysics simulation of a piston operating in high pressure and temperature conditions. Generally, to make such predictions accurate, various details of the geometric and physical models might be accounted for, such as the small geometric features (e.g., holes and grooves), complex material distribution over the geometry, etc. Resolving these details often leads to significant computational time and cost that may turn out to be overkill to predict the desired properties up to an adequate accuracy. The tradeoffs are simply unknown until the model is simulated many times under various assumptions and using different LOGs. 

High-fidelity models (HFMs) for DPM are commonly specified by one or more partial differential equations (PDEs) and initial/boundary conditions (ICs/BCs).  To solve these equations, one often performs a discretization in space over a mesh, e.g., using finite difference \cite{mazumder2015numerical}, finite element  \cite{bathe2006finite}, and finite volume \cite{eymard2000finite} or spectral methods \cite{gottlieb1977numerical}, and selects a time-stepping (e.g., Euler or Runge Kutta \cite{butcher1987numerical}) scheme for integration. The mesh resolution or the number of basis functions are determined by the user, to satisfy accuracy and stability requirements. 
The computational time and cost grows at a quadratic (in 2D) or cubic (in 3D) rate with decreasing mesh length, if not worse \cite{boz2014effects,resolutionrelation2020}. For example, the deformation and temperature fields of the piston in Fig. \ref{fig:piston_comsol} is governed by two coupled PDEs describing the elastodynamics and transient heat transfer. A typical dynamic thermo-elastic simulation for a system like this with geometric details and heterogeneous materials may take anywhere from minutes to hours or days, depending on the mesh resolution, integration time-step, and available computing power. Using such HFM and simulations for every part of a complex engineering system (e.g., an entire automobile) is impractical and in most cases even unnecessary.

A common solution to this problem is to use reduced-order models (ROM) that make compromises in accuracy to achieve a better computational performance. However, it is not easy to determine what kind of geometric or material features may be omitted or simplified to retain the key properties one cares about. For example, decreasing the resolution across the model or ``de-featuring'' the mesh may result in prohibitively large errors, while a much simpler model, perhaps engineered with a careful consideration of what features are important for the predictive outcome, may still capture the essential properties at a small fraction of the HFM simulation cost. The domain insight or validation data for such determinations are not always available. A {\it systematic} and {\it domain-agnostic} approach that can be applied automatically to generate flexible ROMs from a given HFM or data is missing. This paper aims to close that gap by providing such a framework that further provides users with ``knobs'' to adjust cost-accuracy tradeoffs, with guaranteed properties, prior to committing to an LOG for simulation; to customize interpretable structures that can be trained by data; and a combination of both, if applicable.

\begin{figure} 
	\centering
	\includegraphics[width=0.8\linewidth]{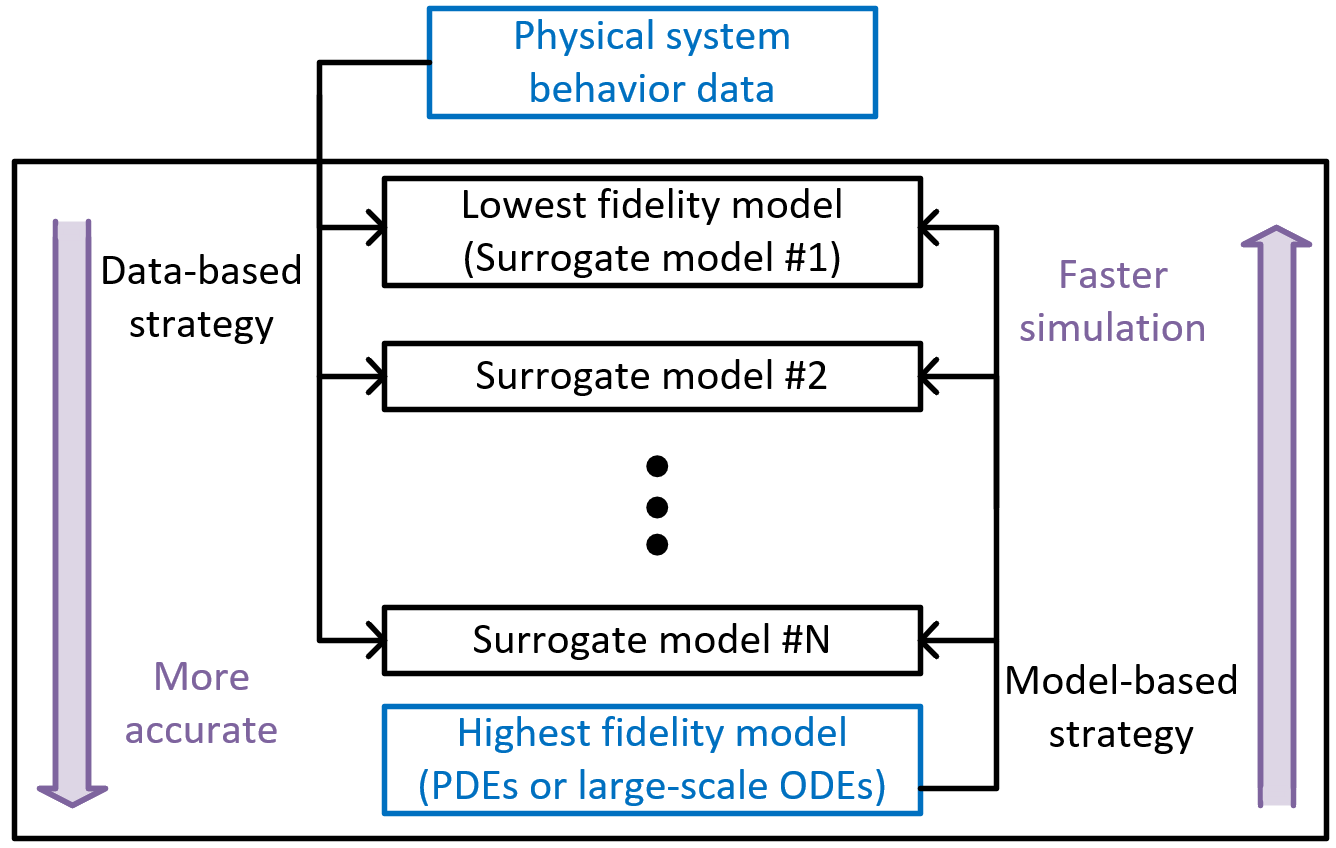}
	\caption{Framework for surrogate modeling}
	\label{fig:scheme}
\end{figure}

We present a framework (Fig. \ref{fig:scheme}) that can systematically generate a family of ROMs for a system that span the cost-accuracy tradeoff spectrum. At the one end of this spectrum, one has the HFM described by a number of PDEs, semi-discretized (i.e., discretized in space but not in time) into a large system of coupled ordinary differential equations (ODEs) or differential algebraic equations (DAEs). One may also have access to data sets representing the behavior, i.e., response of the HFM (or actual physical system) to certain stimuli such as ICs/BCs or source/sink terms.

The generated family of surrogate models may provide one or a combination of properties: (1) flexible cost-accuracy tradeoffs (2) a priori error bounds (3) a priori guarantees of preserving properties that are critical for physical fidelity (4) interpretability. Equipped with this framework, physicists and engineers can quickly choose the LOG at which the ROM can model the system and predict its behaviors most efficiently up to the required accuracy. 

\subsection{Contributions}

This paper has three major contributions:

\begin{enumerate}
    \item A MOR approach from the system and control theory is adapted in developing ROMs for DPMs, using semi-discretization (i.e., discretization in space, but not time) to covert PDEs to a large system of ODEs/DAEs.
    \item A domain-agnostic mechanism is proposed to automatically translate a presupposed topological structure for LPMs to a system of ODEs, which supports both single physics and multiphysics couplings.
    \item A hybrid approach for surrogate modeling is proposed, which, not only maintains all desirable properties of the model-based method, but also retrieves the interpretability of the physical system through imposing LPM structure.
\end{enumerate}

\section{Related work}

\begin{figure*}[!htb]
	\centering
	\includegraphics[width=0.96\linewidth]{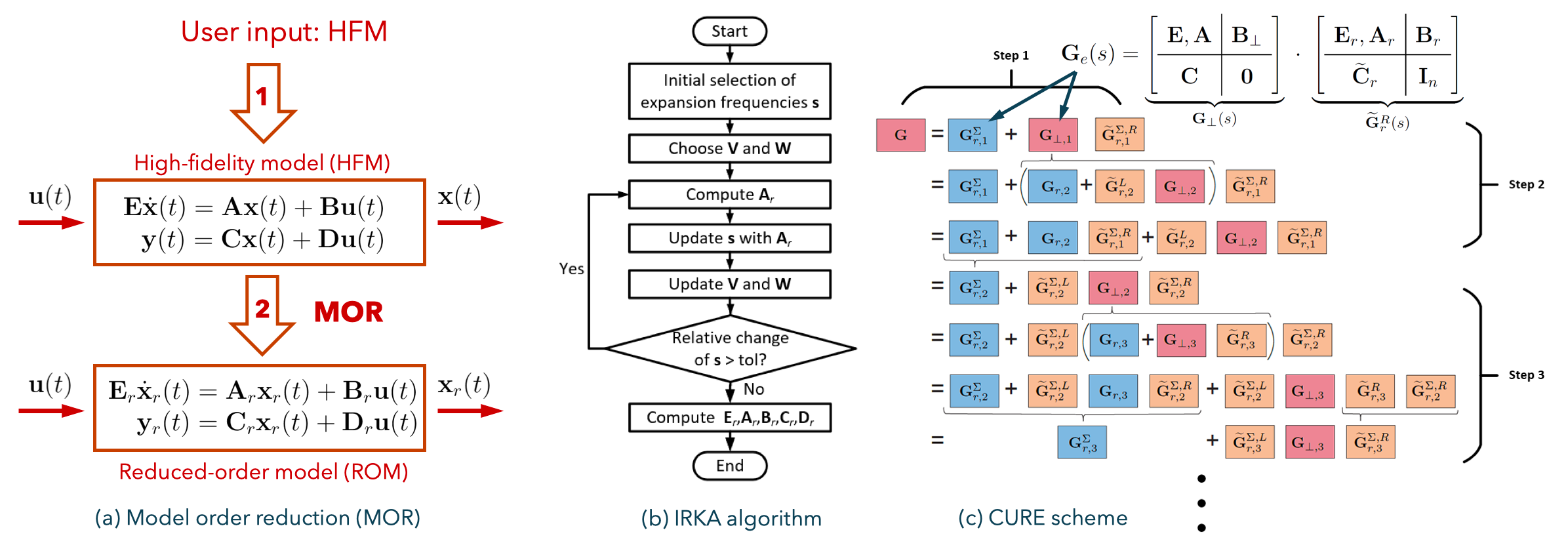}
	\caption{The workflow for the model-based approach to LPM construction. The user provides (a) the HFM, obtained by semi-discretization of a set of PDEs into a large system of ODEs/DAEs (e.g., via finite element spatial discretization). The algorithm re-arranges the second-order equations into twice as many first-order equations in the state-space representation. The state-space representation can be projected to a lower-dimensional space by various MOR techniques, e.g., (b) IRKA. We use an improved version of IRKA, called (c) the CURE scheme. The diagram in (c) is adopted from \cite{panzer2014model} with modification.} \label{fig_model_based}
\end{figure*}

Different approaches to surrogate modeling of physical systems can be broadly categorized into model-based and data-driven methods. The model-based methods such as MOR are generally used when a ``white box'' HFM is available. Bottom-up MOR techniques aim at finding a good approximation of the HFM such that the approximation is numerically efficient and stable while preserving certain physical and numerical properties of the HFM \cite{lohmann2000introduction,baur2014model}. The MOR process usually starts with a large system of ODEs and produces a significantly smaller system of ODEs that is guaranteed to predict a similar solution for given stimulus. Fig. \ref{fig_model_based} (a) illustrates the key ideas in MOR. By contrast, the data-driven methods, including machine learning (ML) or regression approaches to parameter estimation for a small system of ODEs, are generally used if the HFM is a ``black box'', where some input and output data sets are given, but the model's inner working mechanism is not available. 

The commonly-used MOR methods for linear time-invariant (LTI) systems are the balanced truncation (BT) method  \cite{chahlaoui2005model,reis2008survey,stykel2002analysis} and the rational Krylov subspace (RKS) method \cite{bai2002krylov,bai2005dimension,freund2004sprim,grimme1997krylov}, which are based on projection. The principle of the BT method is to remove the states that have weak controllability and observability \cite{chahlaoui2005model}. The BT method is not time-efficient for large-scale models that have more than a few thousands of variables because finding out such states is a time-consuming process. By contrast, the RKS method matches several most significant terms of the Tayler series expansion of the ROM transfer function, expanded around carefully selected frequencies, to those of the HFM. Although RKS is more scalable than BT, it has several drawbacks; for example, it does not guarantee preservation of the HFM's stability (i.e., even if the HFM is stable, the ROM may not be) and the Taylor series expansion frequencies have to be selected manually, which is nontrivial. 

To overcome the above limitations of RKS, researchers have developed an improved method based on an iterative algorithm \cite{gugercin2008h_2}, which allows adaptively choosing the expansion frequencies using the mirror image of the poles of the obtained transfer function of ROM. This algorithm is considered as a ``gold standard'' among the projection-based MOR methods that minimize the norm of approximation error between ROM and HFM transfer functions for a given target model order\footnote{The ``order'' of a model is the number of state variables, and the number of first-order ODEs, in the state-space representation.}. Compared to RKS, the IRKA has the advantage of the ability to automatically choose expansion frequencies by iteration while maintaining the model stability \cite{gugercin2008h_2}. The algorithm of IRKA is summarized in Fig. \ref{fig_model_based} (b). However, it can neither guarantee monotonically decreasing errors with every iteration, nor ensure that the error converges to a local minimum \cite{beattie2009trust}. The CUmulative REduction (CURE) scheme \cite{panzer2014model} was later developed to improve several critical properties of the IRKA. The algorithm of the CURE scheme can be found in \cite{panzer2014model} and is repeated in Fig. \ref{fig_model_based} (c). Particularly, to maintain the model stability, a stability-preserving, adaptive rational Krylov (SPARK) algorithm was developed \cite{panzer2014model}, which is usually embedded in the CURE scheme to generate a family of stable ROMs with increasing model orders by sequential accumulation in a single MOR process. Both IRKA and SPARK share the same model-reduction principle as the RKS methods and are both numerically efficient. However, compared to the IRKA method, the SPARK+CURE has further advantages such as ensuring stability, monotonic convergence, and a priori error guarantees, as well as automatic model order decision making capabilities. 

Table \ref{fig:MOR-comparison} illustrates a comparison of the properties of the four MOR methods mentioned above. The major drawback of all of these projection-based MOR techniques is the loss of physics-based structure and interpretability. There exist structure-preserving (e.g., symplectic) MOR approaches \cite{peng2016symplectic} that can remedy this shortcoming; however, the added interpretability may come at the expense of losing the desirable properties in Table \ref{fig:MOR-comparison}. We will present a simple hybridization strategy to augment CURE with a data-driven approach to retrieve the physical structure with a small compromise in the a priori error guarantees.

\begin{table}[!htb]
  \caption{Comparison of properties of four commonly-used MOR methods (CURE is uniquely positioned)}
  \label{fig:MOR-comparison}
  \includegraphics[width=\linewidth]{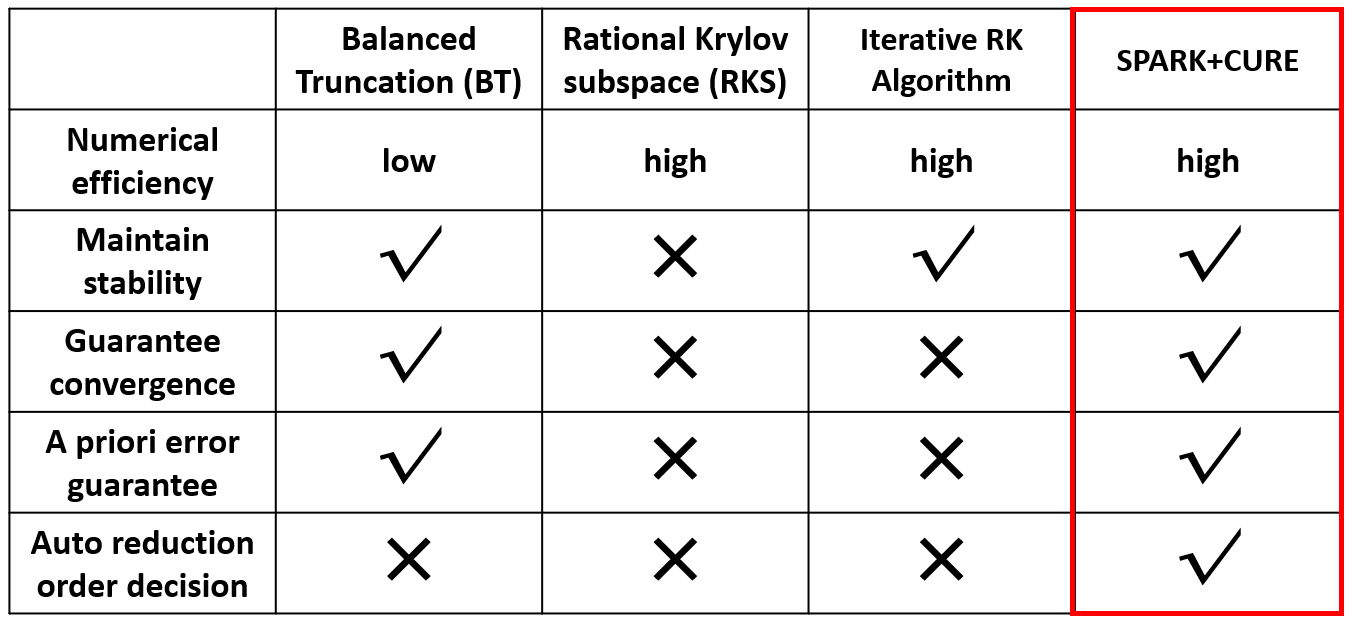}
\end{table}

Popular data-driven methods include, but are not limited to, symbolic regression \cite{schmidt2009distilling,schmidt2011automated}, recurrent neural networks \cite{sussillo2009generating}, evolved regulatory networks \cite{franccois2007deriving}, and physics-informed neural networks \cite{raissi2019physics}. These methods have demonstrated the ability to accurately replicate HFM after sufficient training and testing. Their development often requires domain-specific insight to select the proper set of differential equations and directly build these equations into a constrained learning structure and/or penalize the loss function by residual errors. In this paper, however, we provide an approach that only requires the user insight in choosing an appropriate LPM topology while the system equations can be automatically generated in a domain-agnostic fashion.

\section{Methods and Algorithms}
Below, we introduce our model-based and data-driven approaches to bottom-up MOR of white box systems and top-down surrogate modeling of black box systems, respectively. At the end, we provide guidelines on how to use both approaches in a hybrid setting for ``gray box'' systems.

\subsection{A Model-Based Approach}

For the model-based approach, we use the SPARK+CURE \cite{panzer2014model} due to its rigorous mathematical underpinnings that lead to strong guarantees for numerical simulation. In a nutshell, this method has several important properties, namely: (1) automatic discovery of proper expansion frequencies; (2) guaranteed preservation of stability; (3) guaranteed $\mathcal{H}_2-$error convergence; and (4) an a priori $\mathcal{H}_2-$error bound.\footnote{The $\mathcal{H}_2-$error is defined as the $L_2-$norm of the error transfer function over the imaginary line in the frequency domain.}
More specifically, the method not only allows adaptively choosing the expansion frequencies using the poles of the obtained ROM (e.g., similarly to IRKA), but also enables incrementally increasing the model order by cumulatively updating the ROM error transfer function. Particularly, the error of the obtained ROM monotonically converges to zero (i.e., ROM converges to HFM) in the accumulation process. The SPARK+CURE scheme can also receive a tolerance for error, based on which a termination criterion for cumulative reduction can be defined. In other words, the user provides not only the HFM matrices but also a maximal value of error that can be overlooked, and the algorithm generates the lowest-order ROM (hence the fastest to simulate) that is still guaranteed to remain within the error tolerance, without actually performing any numerical simulation.

\subsection{A Data-Driven Approach}

\begin{figure*}
	\centering
	\includegraphics[width=0.96\linewidth]{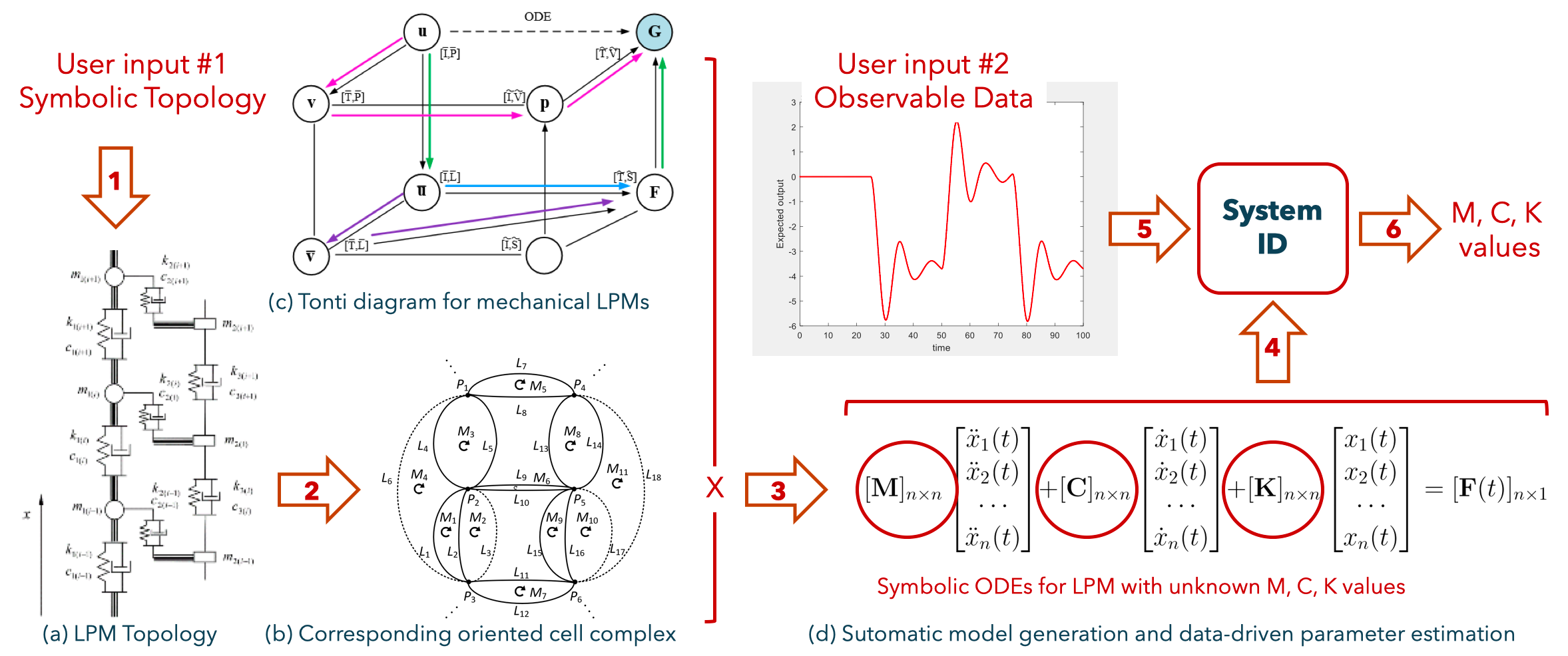}
	\caption{The workflow for the data-driven approach to LPM construction. The user provides (a) a topology for the LPM (i.e., symbolic network of inter-connected components), which is then converted to (b) the common language of abstract oriented cell complexes \cite{wang2019topological}. (c) The Tonti diagram \cite{tonti2013mathematical} converts the cell complex representation to (d) a system of symbolic ODEs with unknown constitutive parameters. The parameters are learned from data using standard system identification techniques.} \label{fig_data_driven}
\end{figure*}

In a recent article \cite{wang2019topological}, common reference semantics for lumped parameter system modeling were presented, based on algebraic topological foundations of network theory \cite{roth1955application,branin1966algebraic}, which can serve as a unifying abstraction of system modeling languages such as \textsf{Modelica} \cite{elmqvist1997introduction}, \textsf{Simulink} \cite{chaturvedi2009modeling}, linear graphs \cite{rowell1997system}, and bond graphs \cite{paynter1961analysis}, etc. A key advantage of using this abstraction is the ability to automatically map a given topological structure for the LPM (e.g., a circuit graph or mass/spring/damper network) to a set of governing ODEs. These ODEs have built-in conservation laws in the LPM context such as Kirchhoff's current and voltage laws for electrical and thermal circuits, superposition of forces and Newton's laws of motion in multi-body dynamics, and so on. They also include constitutive laws associated with lumped components such as springs and dampers in mechanical systems, resistors and capacitors in analog circuits, conductors in heat transfer, and so on. The recipe for generating the ODEs from system topology in \cite{wang2019topological} is given by Tonti diagrams \cite{tonti2013mathematical} of network theory (Fig. \ref{fig_data_driven} (c)). The Tonti diagram is a composition of topological and algebraic operators that map data associated with different cells in an oriented cell complex representation of the LPM network; for instance, in an electrical circuit, the superposition of incoming/outgoing currents on incident wires (i.e., $1-$cells) to a junction (i.e., $0-$cells) is captured by a boundary operator (from $1-$cells to $0-$cells), whereas resistance, capacitance, inductance, and other constitutive relations are in-place algebraic relations that keep data on $1-$cells. These operators are represented by different types of arrows on the Tonti diagram. The key advatange of using this approach to equation generation is its generalizability to various domains of physics and possible multi-physics, as the underlying  topological and algebraic operations are common to mechanical, electrical, thermal, and other systems \cite{tonti2013mathematical}.

Constitutive relationships in LPM capture ``effective'' phenomenological properties of the system at a certain LOG of choice, unlike the constitutive relations in DPM derived directly from well-documented material properties. Except in cases where an LPM is directly generated from a system model with modular components (e.g., actual springs/dampers in an automobile suspension assembly), the parameters for the artificial LPM components are not easily obtained from geometric and material properties. These parameters must be estimated from data by solving an optimization problem (e.g., least squares regression).

Figure \ref{fig_data_driven} illustrates the workflow for the data-driven LPM construction.
Given an experimental or simulation data set and an LPM topology in Fig. \ref{fig_data_driven} (a), we first convert the LPM from the domain-specific format (e.g., \textsf{Modelica}) to the domain-agnostic canonical form (i.e., oriented cell complex) as shown in Fig. \ref{fig_data_driven} (b), using the semantics provided in \cite{wang2019topological}. Each $1-$cell is associated with a symbolic constitutive relation and given an initial value to the constitutive parameter. After selecting a state variable of interest, the state equations (i.e., system of second-order ODEs/DAEs) are generated by tracing groups of paths along the Tonti diagram of network theory with the appropriate physical types \cite{wang2019topological}. There are a total of 8 different options for state variables, each of which corresponds to a different groups of paths \cite{wang2019topological}. Once the system of ODEs/DAEs are assembled, we use ordinary least squares regression---although other objective functions and optimization techniques are certainly applicable---to iteratively update the constitutive parameters until the solution of the state equation fits the given data. 

In a later section, we will apply this method to obtain an LPM for a mechanical problem (single physics) and a thermo-mechanical problem (coupled multiphysics).

\subsection{A Hybrid Approach}

Both of the model-based and data-driven approached mentioned above have pros and cons; in particular:
\begin{itemize}
    \item The model-based approach has the advantage of starting from first principles and requiring no data other than basic material properties that are used in the HFM (e.g., constitutive part of the PDEs).
    \item The model-based method also provides rigorous guarantees that are rarely avialble in data-driven methods; however, the resulting ROMs are often not interpretable.
    \item The data-driven method provides a mechanism to specify the desired LPM with interpretable constitutive relations (``artificial'' components as $1-$cells in the cell complex) if the user has insight to choose an appropriate structure.
\end{itemize}

In particular, even though the SPARK+CURE method has several useful properties, it has the following limitations: (1) the HFM itself must be dissipative (hence stable) to begin with (2) the resulting HFM is a system of first-order ODEs/DAEs with dense state-space matrices which may not be refactored into a second-order system (e.g., with mass, spring, and damper matrices in mechanical LPM) for component-wise interpretability and (3) the rigorous guarantees are only valid for LTI systems, although there are several ways in which it can be generalized to nonlinear systems, some of which are ongoing research.

On the other hand, the regression-based system identification for data-driven fitting becomes impractical when the HFM is too costly to simulate (to obtain synthetic data) and the experimental measurements of adequate quality are not available. One can always use as many data points as one can obtain within the computational and experimental budget, but the lack of guarantees in predicting out-of-training inputs undermines the ROM's reliability.

To remedy the shortcomings of the the model-based and data-driven approaches, we developed a ``hyrbid'' approach to achieve the best of both worlds. For each generated ROM $\mathsf{M}$ from the SPARK+CURE method, we use this ROM to produce training data for another surrogate LPM $\mathsf{M}'$ of the same order $r \ll n$ via system identification. Although the training data is itself erroneous, its $\mathcal{H}_2-$error is bounded by the CURE framework (denoted by $\varepsilon_\mathsf{M}$). The error between the two ROMs (denoted by $\bar{\varepsilon}_{\mathrm{rel}}$), on the other hand, can be computed by solving two small-scale algebraic Lyapunov equations.%
\footnote{Note that solving such a system for the full-order HFM would be almost as prohibitive as numerical simulation.}
Hence, the $\mathcal{H}_2-$error for the surrogate LPM can also be guaranteed via a triangular inequality:
\begin{equation} \label{Eq：totalH2Bound}
    \bar{\varepsilon}_{\mathsf{M}'} \le \bar{\varepsilon}_{\mathsf{M}} + \bar{\varepsilon}_{\mathrm{rel}}.
\end{equation}

\section{Applications and Results}

\subsection{Spatial Discretization}
To generate data for our data-driven method by using the ROM generated from the SPARK+CURE method, we will use LTI models and FEA discretization \cite{bathe2006finite}, but virtually every spatial discretization works as long as it generates a system of LTI ODEs/DAEs. Note that we do not perform temporal discretization for MOR, as the SPARK+CURE method is best described in terms of continuous and differentiable functions in time. Temporal discretization, on the other hand, comes into play for numerical simulation of the system of ODEs/DAEs (before or after MOR) by finite time-stepping to approximate integration with given ICs.

\subsection{Mechanical and Thermal Examples}
We will generate data using two geometric domains; namely, a simple cylinder (Fig. \ref{fig:ex-AT-ME}) to illustrate the basic ideas with a closed-form ground truth to compare against, and a piston assembly (Fig. \ref{fig:PistonMesh}) to demonstrate a slightly more complicated case involving one-way coupled thermo-elasticity. In both examples, we assume linear elastic material properties undergoing dynamic mechanical and/or thermal loads. The BCs in these examples are fixed displacement at some surfaces and uniform pressure and heat fluxes at other surfaces. 
In principle, the approach can be  applied to domains of arbitrary shapes, material distributions, ICs/BCs, and excitations. More extensive testing with strongly (e.g., two-way) coupled mutliphysics problems is required to further validate the approach, which is suitable for a full paper.

\subsection{Preliminary Results}

\begin{figure} [ht!]
	\begin{subfigure}{.11\textwidth}
		\centering
		\includegraphics[width=0.60\linewidth]{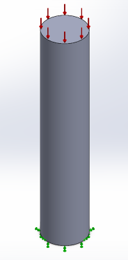}
		\caption{Cylinder}  
		\label{fig:Cylinder}
	\end{subfigure}%
	\begin{subfigure}{.11\textwidth}
		\centering
		\includegraphics[width=0.4\linewidth]{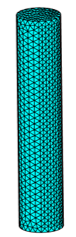}
		\caption{Mesh}
		\label{fig:Cylinder-mesh}
	\end{subfigure}
	\begin{subfigure}{.12\textwidth}
		\centering
		\includegraphics[width=\linewidth]{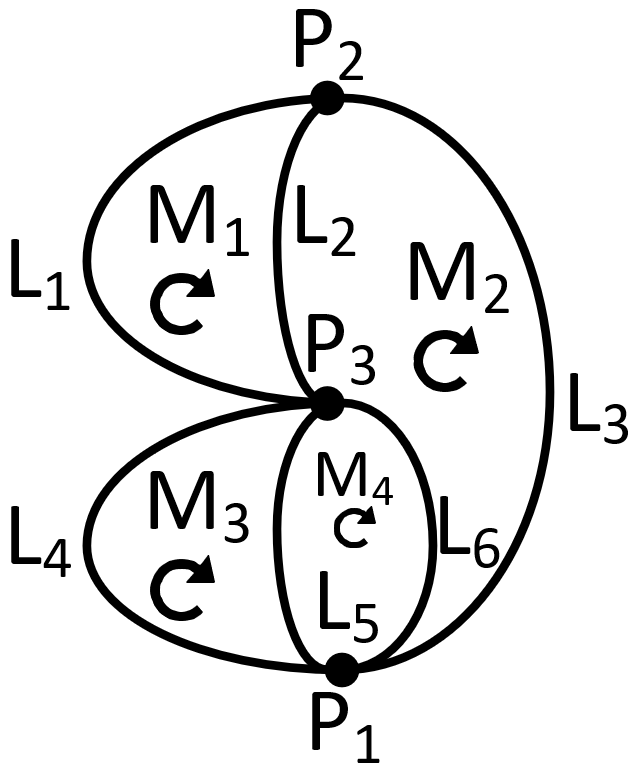}
		\caption{Topology}  
		\label{fig:ex-AT2}
	\end{subfigure}
	\begin{subfigure}{.12\textwidth}
		\centering
		\includegraphics[width=.9\linewidth]{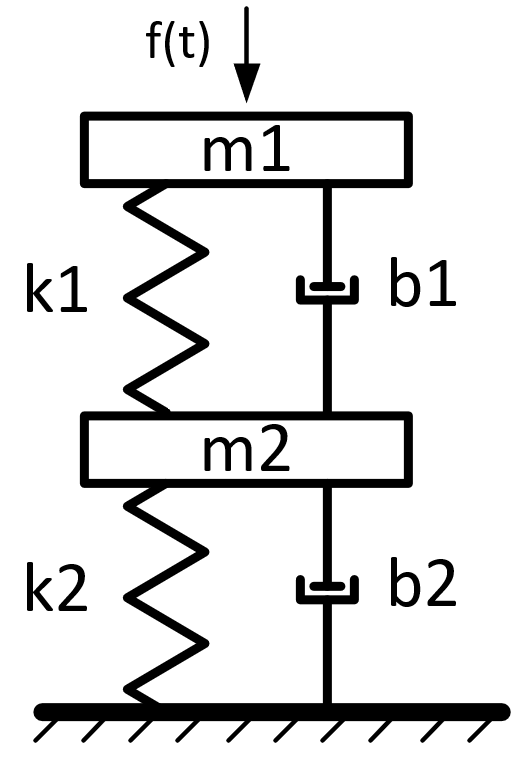}
		\caption{System}  
		\label{fig:LPM_MKC}
	\end{subfigure}
	\begin{subfigure}{.5\textwidth}
		\centering
		\includegraphics[width=0.9\linewidth]{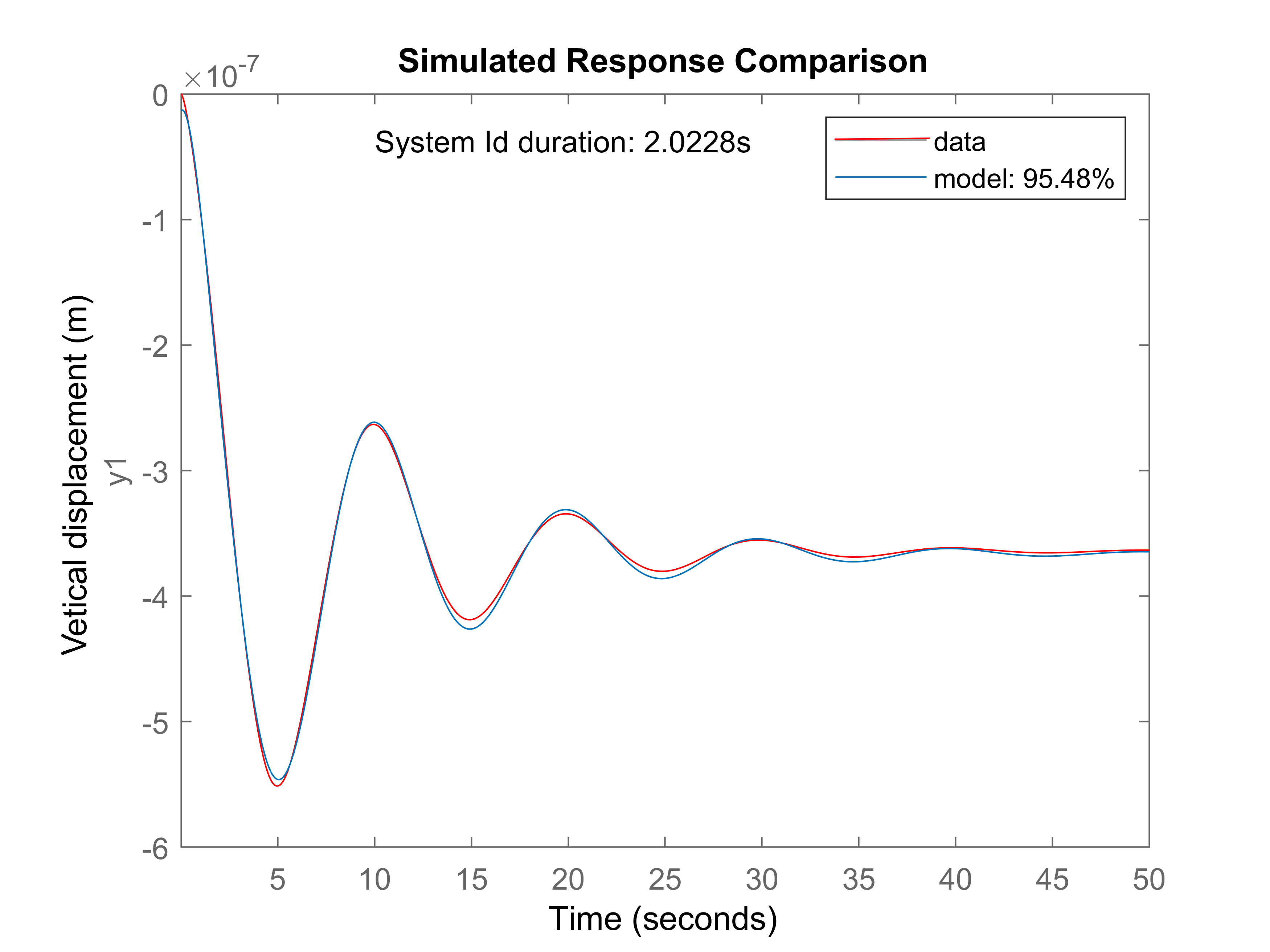}
		\caption{Simulation results comparison}
		\label{fig:ex-fit-ME2}
	\end{subfigure}
	\caption{The topological structure of the physical space of a lumped mass-spring-damper system and the comparison of simulation results between the HFM, the reduced FEA model, and the optimal model}
	\label{fig:ex-AT-ME}
\end{figure}

\begin{figure} [ht!]
	\begin{subfigure}{.5\textwidth}
		\centering
		\includegraphics[width=0.9\linewidth]{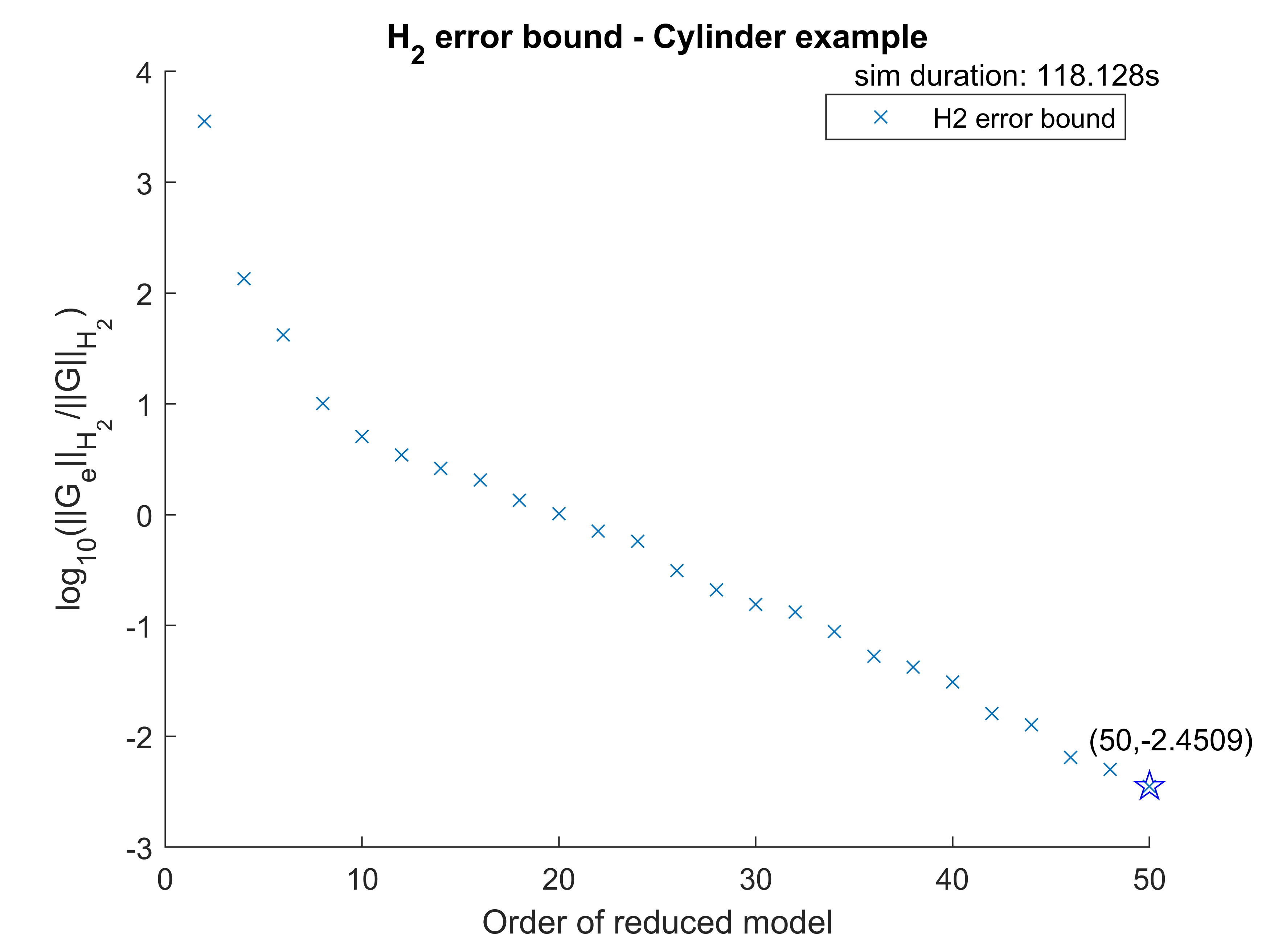}
		\caption{A priori relative $\mathcal{H}_2-$error bound}
		\label{fig:error_bound_case9}
	\end{subfigure}
	\begin{subfigure}{.5\textwidth}
		\centering
		\includegraphics[width=0.9\linewidth]{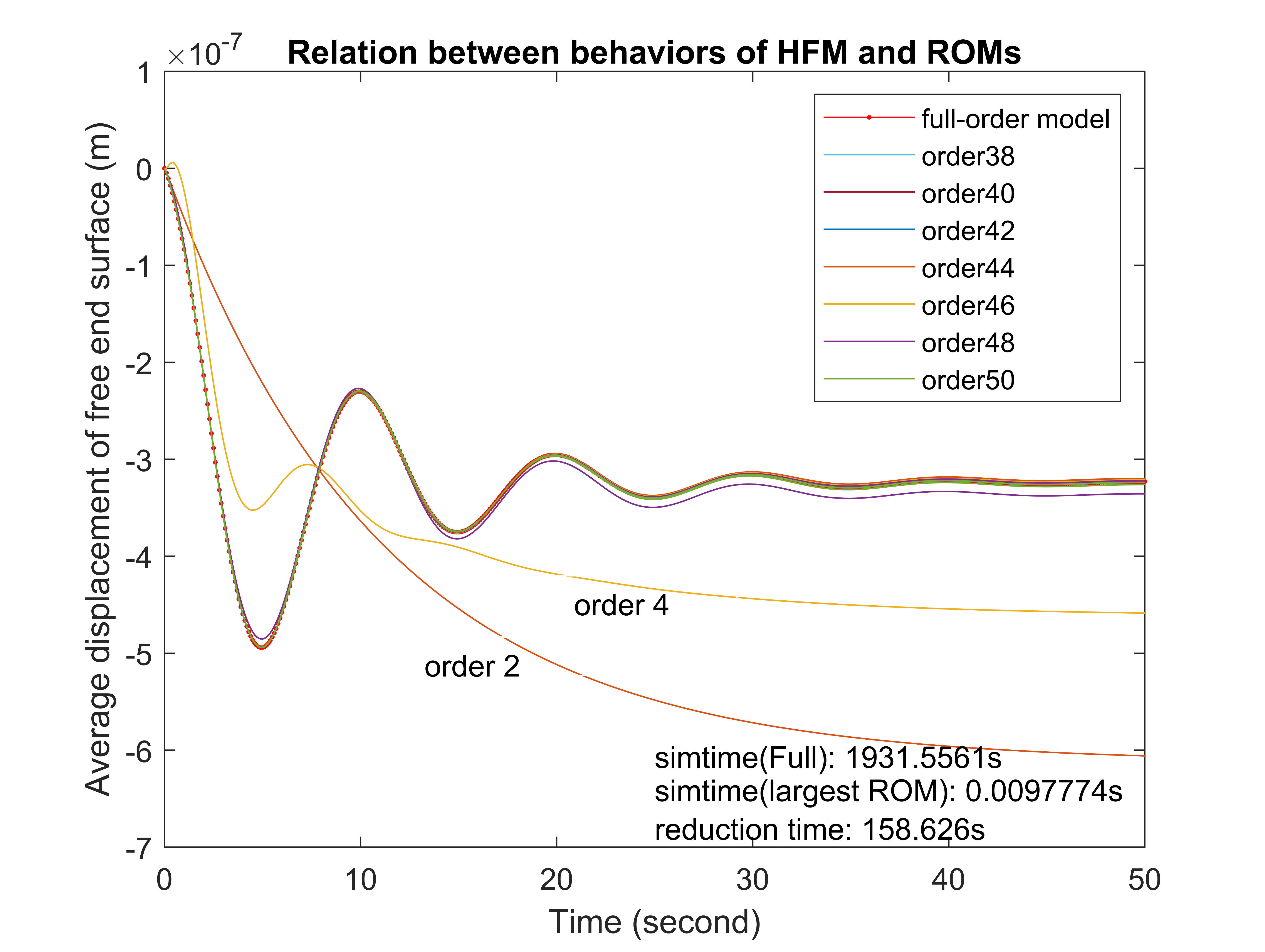}
		\caption{Simulation results comparison}  
		\label{fig:behavior_case9}
	\end{subfigure}
	\caption{Simulation results comparison between HFM and a family of ROMs, and a priori relative $\mathcal{H}_2-$error bounds.}
	\label{fig:CURE_MOR_cylinder}
\end{figure}

Consider the homogeneous cylinder in Fig. \ref{fig:ex-AT-ME} (a) with one end fixed to the ground and the other end bearing a constant unit pressure. The cylinder is discretized using second-order tetrahedral finite elements, leading to 17,430 nodal displacement variables. Using the vertical average displacement of the top surface of the cylinder as the solution of interest, we use the SPARK+CURE to generate a family of ROMs. Figure \ref{fig:CURE_MOR_cylinder} (a) shows the a priori relative $\mathcal{H}_2-$error bounds for each of them. It can be observed that the error bound monotonically decreases with increasing order of the ROMs. For visualization, we compare the simulation results of the ROM and the original HFM (finite elements simulation) as shown in Fig. \ref{fig:CURE_MOR_cylinder} (b). 
We select the ROM of order $r=50$ to generate simulation data, whose a priori relative $\mathcal{H}_2-$error bound is $10^{-2.4509} \approx 0.35\%$ as shown in Fig. \ref{fig:CURE_MOR_cylinder} (a).

\begin{figure} [ht!]
	\begin{subfigure}{.24\textwidth}
		\centering
		\includegraphics[width=\linewidth]{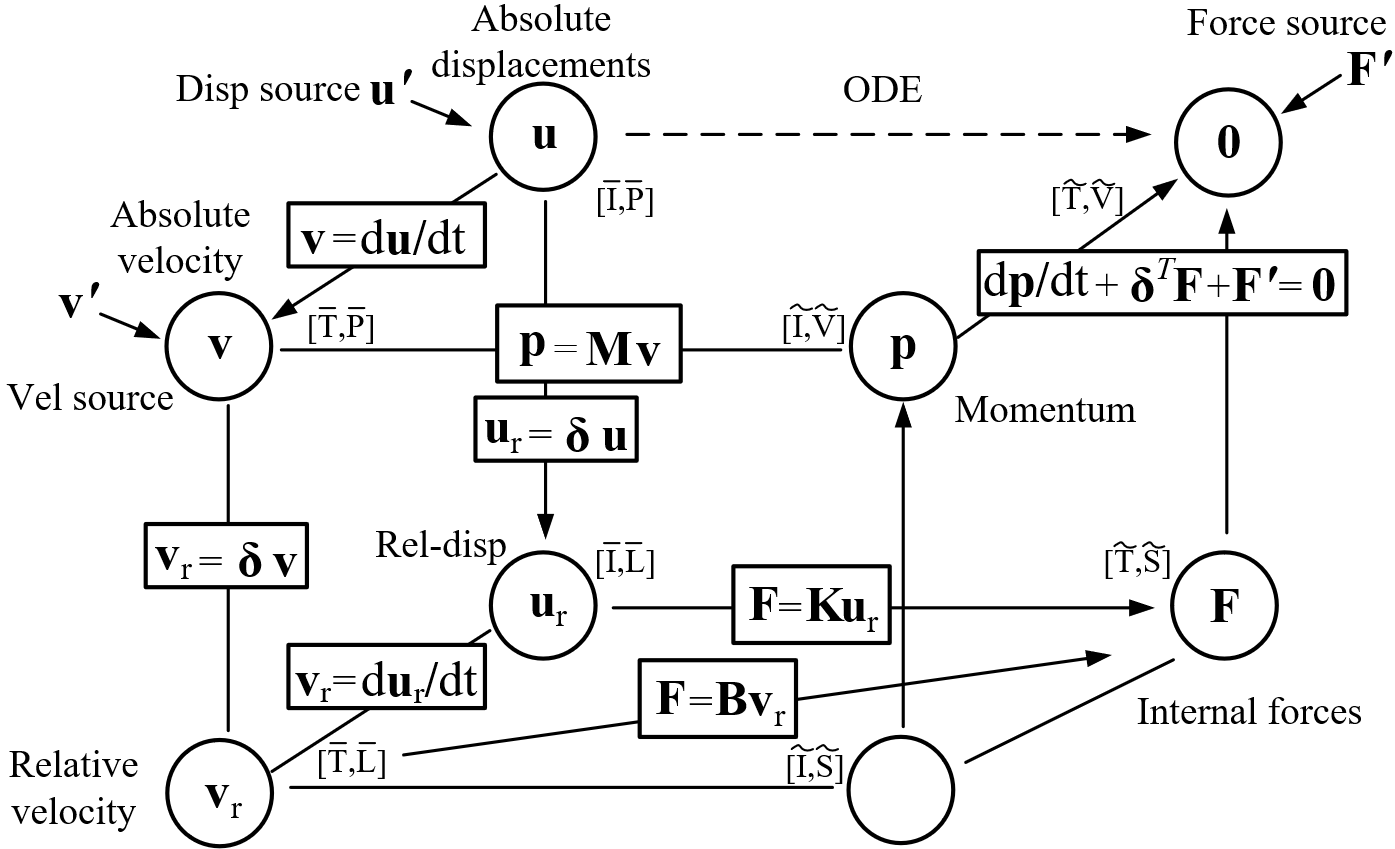}
		\caption{Tonti diagram}  
		\label{fig:TD_ME}
	\end{subfigure}%
	\begin{subfigure}{.24\textwidth}
		\centering
		\includegraphics[width=\linewidth]{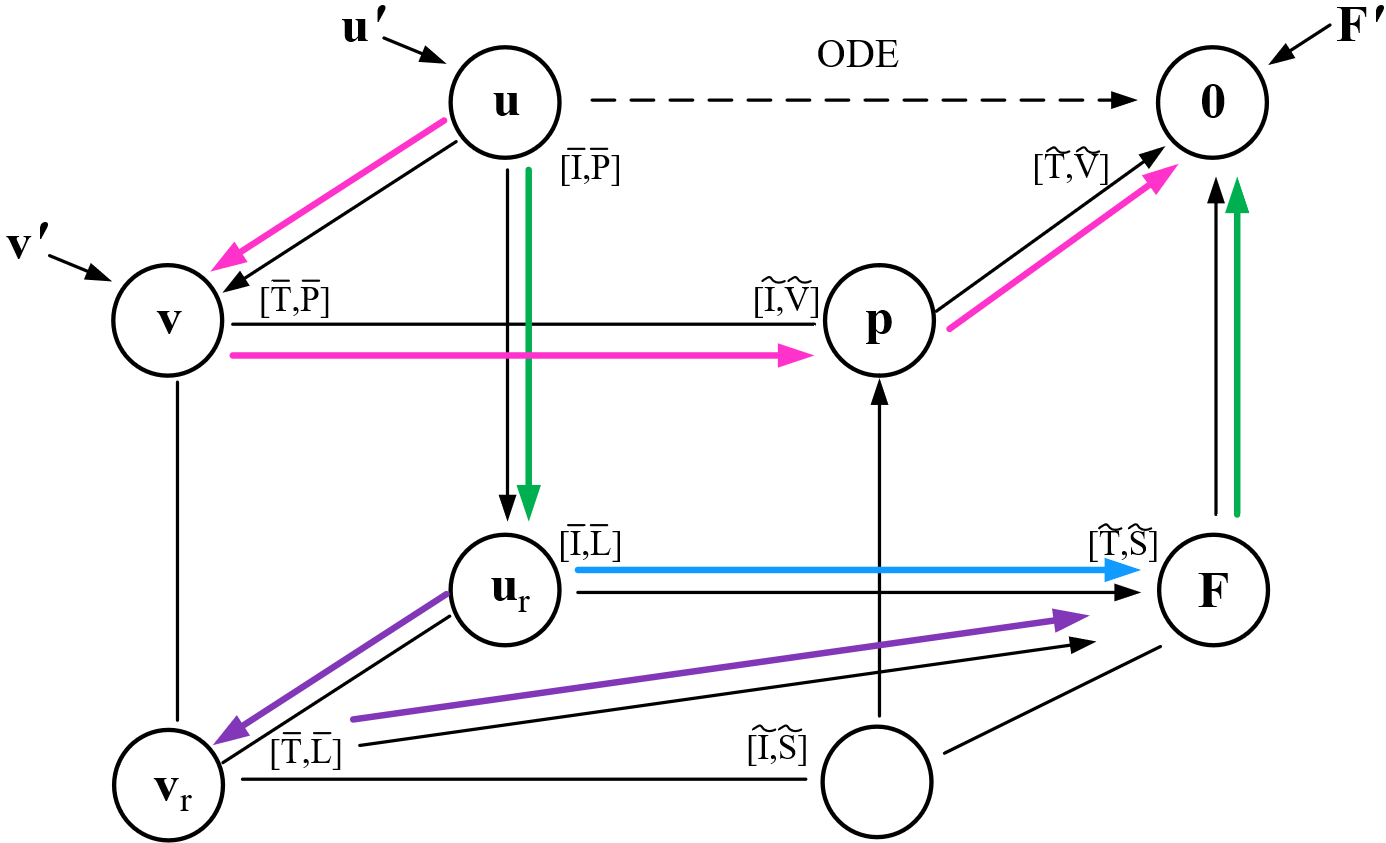}
		\caption{Paths}
		\label{fig:TD_ME_paths}
	\end{subfigure}
	\caption{Tonti diagram of lumped mass-spring-damper system (with lumped sources) and paths for generating governing equations from LPM topology.}
	\label{fig:TD_ME|paths}
\end{figure}

To find an interpretable ROM with a desired topological structure, we initialize an oriented cell complex (Fig. \ref{fig:ex-AT-ME} (c)) representing  a lumped mass-spring-damper network with 2 degrees of freedom (Fig. \ref{fig:ex-AT-ME} (d)). We label the $1-$cells with $L_3$, $L_6$ for masses, $L_2$, $L_5$ for springs, and $L_1$, $L_4$ for dampers. The Tonti diagram of mechanical LPM is shown in Fig. \ref{fig:TD_ME}, and the paths traced to generate the system of ODEs are shown in Fig. \ref{fig:TD_ME|paths} (b). We apply an ordinary least squares regression \cite{miller2006method} to find the values of the lumped masses, stiffness coefficients, and damping coefficients and compare the simulation result for the optimal LPM against the ROM and the HFM (Fig. \ref{fig:ex-AT-ME} (e)). It can be observed that the simulation results between the ROM and the HRM are close and the simulation results between the optimal LPM and the ROM match well, with a normalized root-mean-square error 
(NRMSE)\footnote{The NRMSE is defined as the $L_2-$norm of the error signal (between ROM and LPM) in the time domain, normalized by the variation interval of the ROM.} of 4.52\%.
The relative $\mathcal{H}_2-$error bound computed from (\ref{Eq：totalH2Bound}) against HFM is 0.184.

\begin{figure} [ht!]
	\begin{subfigure}{.24\textwidth}
		\centering
		\includegraphics[width=0.6\linewidth]{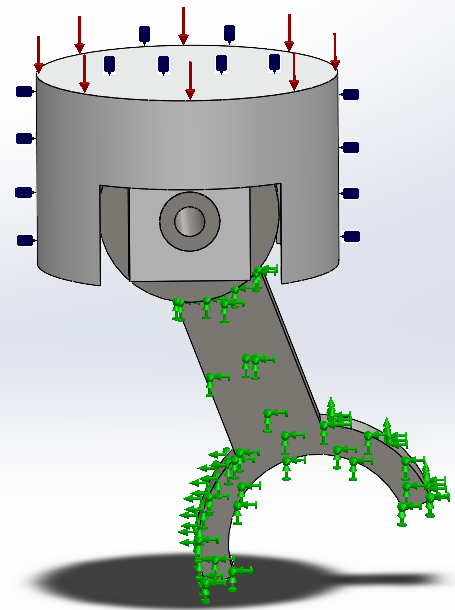}
		\caption{Piston}  
		\label{fig:piston_coupled}
	\end{subfigure}%
	\begin{subfigure}{.24\textwidth}
		\centering
		\includegraphics[width=0.6\linewidth]{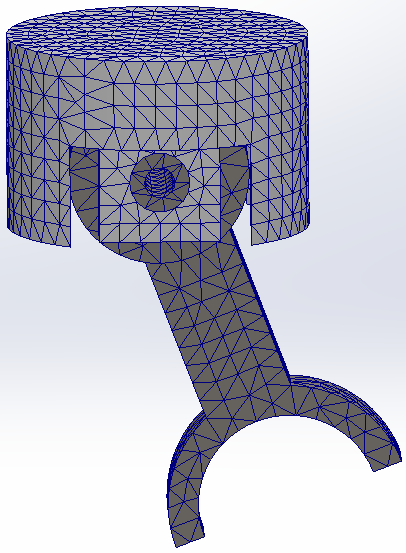}
		\caption{Mesh of piston}
		\label{fig:piston_mesh}
	\end{subfigure}
	\caption{A Piston and its FEA mesh for HFM simulation.}
	\label{fig:PistonMesh}
\end{figure}

Next, we apply the approach to a slightly more complex problem with weak thermo-elastic coupling, over a piston geometry shown in Fig. \ref{fig:PistonMesh} (a), in which the temperature change has an impact on structural field, not vice versa. Parts of the piston are combined into one single part to avoid relative motion. The piston bears a unit pressure (red arrows) on the top surface, a constant heat flux (blue pins) on the piston head, and the crank is fixed (green pins). The solution of interest is the vertical average displacement of the top surface. The data is obtained by simulating a ROM generated from the SPARK+CURE method, with an a priori relative $\mathcal{H}_2-$error bound of 0.0031, computed from (\ref{Eq：totalH2Bound}).

\begin{figure} [ht!]
	\begin{subfigure}{.24\textwidth}
		\centering
		\includegraphics[width=\linewidth]{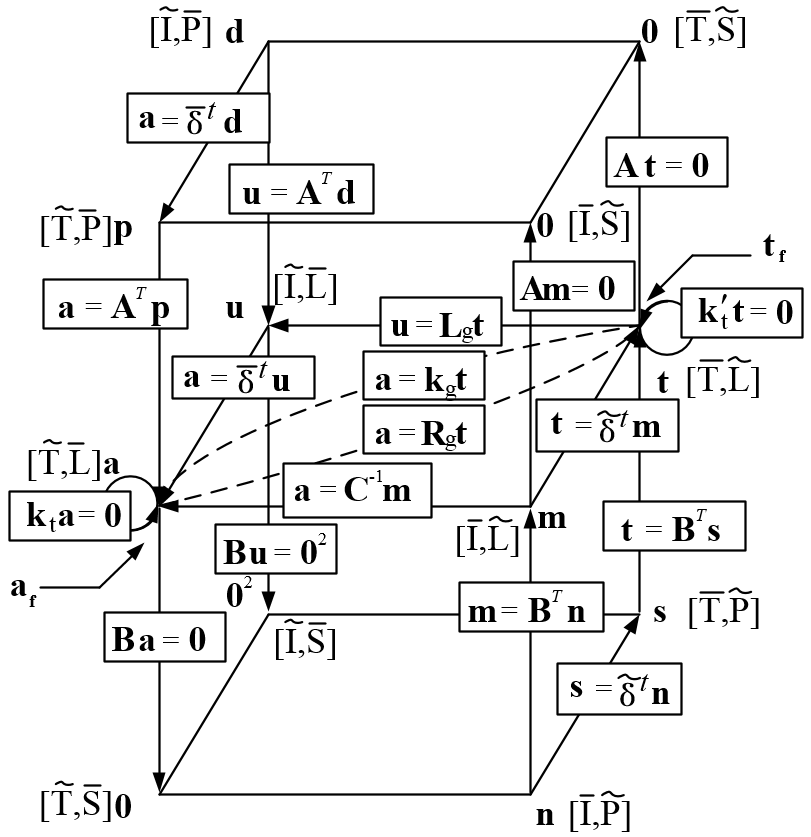}
		\caption{Generalized Tonti diagram}  
		\label{fig:TD_generalized}
	\end{subfigure}%
	\begin{subfigure}{.24\textwidth}
		\centering
		\includegraphics[width=\linewidth]{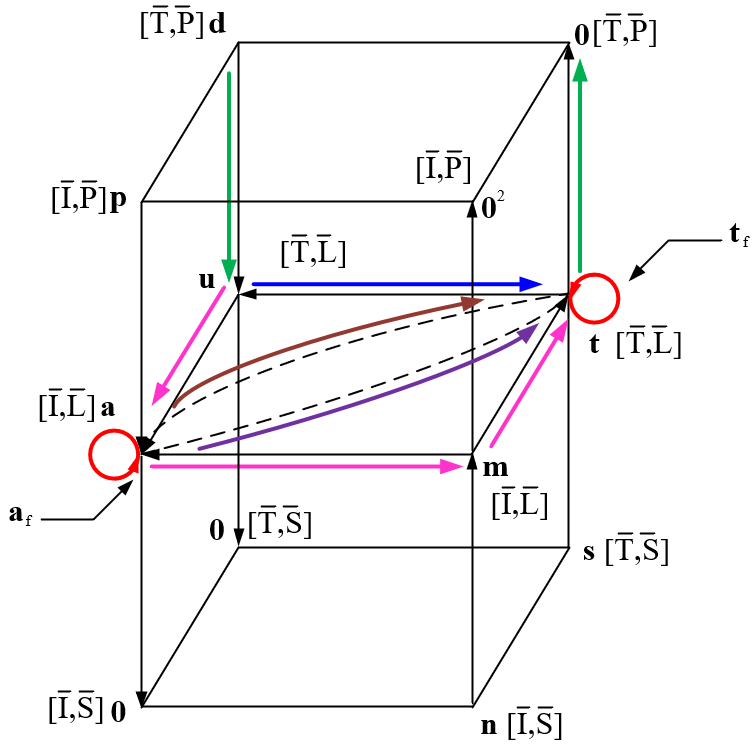}
		\caption{Paths}
		\label{fig:TD_generalized_paths}
	\end{subfigure}
	\caption{Generalized Tonti diagram of multi-domain LPM and paths for generating governing equations.}
	\label{fig:TD_generalized|paths}
\end{figure}

\begin{figure} [ht!]
	\centering\includegraphics[width=\linewidth]{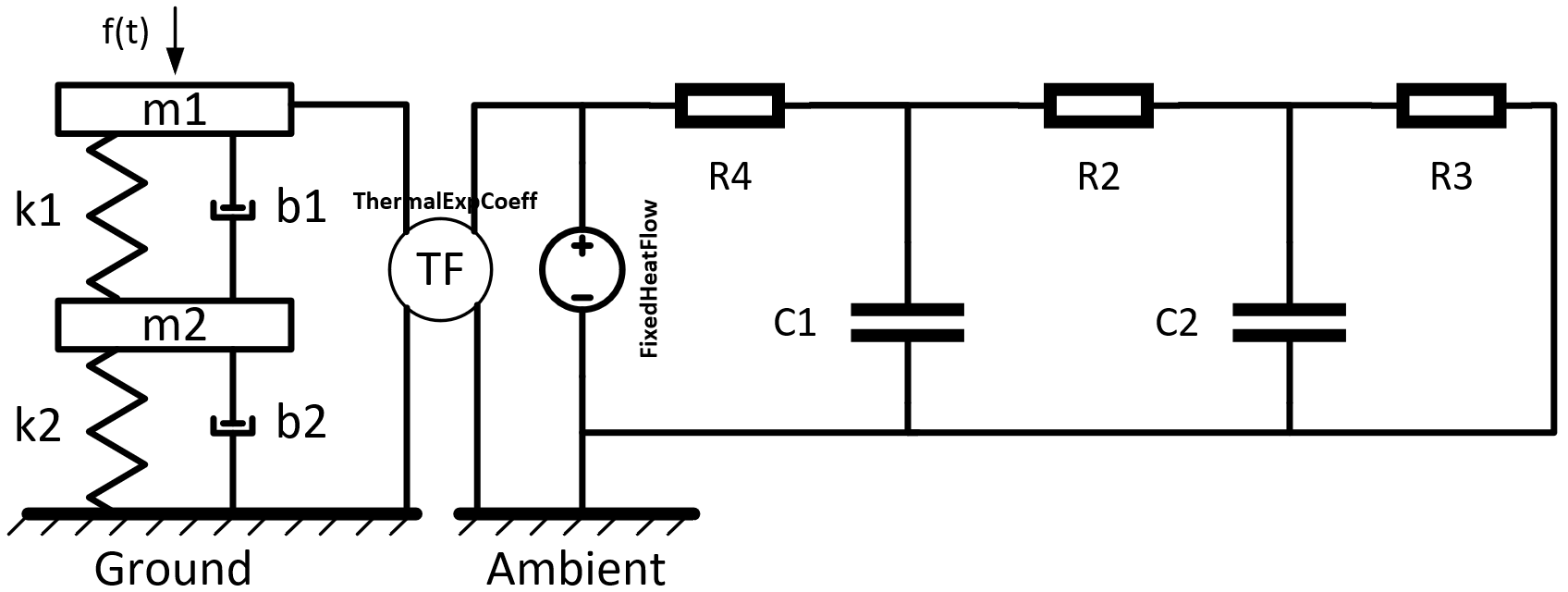}
	\caption{A thermo-mechanical LPM with a TF.}
	\label{fig:LPM_coupled}
\end{figure}

\begin{figure} [ht!]
	\centering\includegraphics[width=0.66\linewidth]{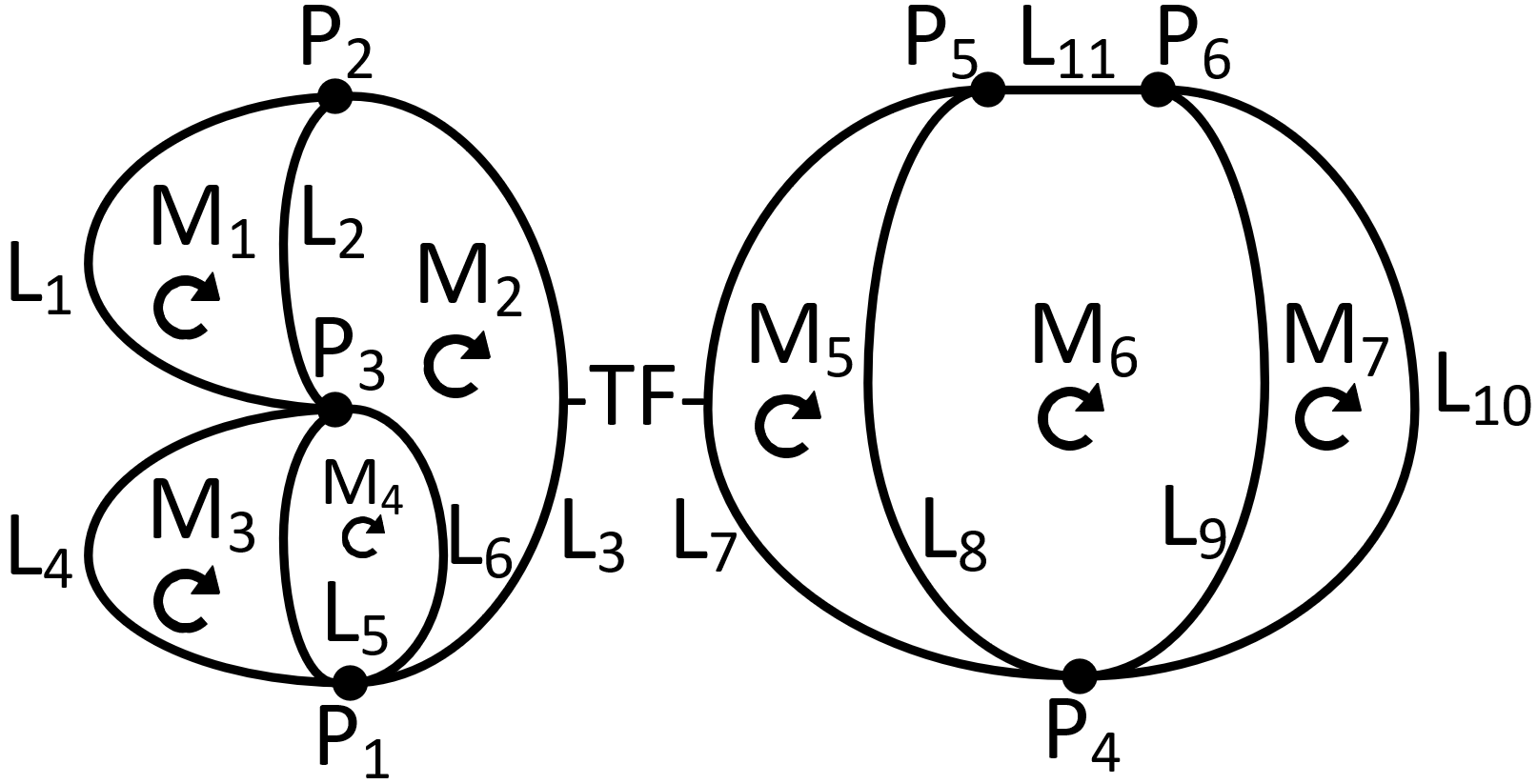}
	\caption{The topological structure of the physical space of a thermo-mechanical LPM with a transformer (TF) coupling the two physics.}
	\label{fig:ex-AT3}
\end{figure} 

\begin{figure} [ht!]
	\centering\includegraphics[width=0.9\linewidth]{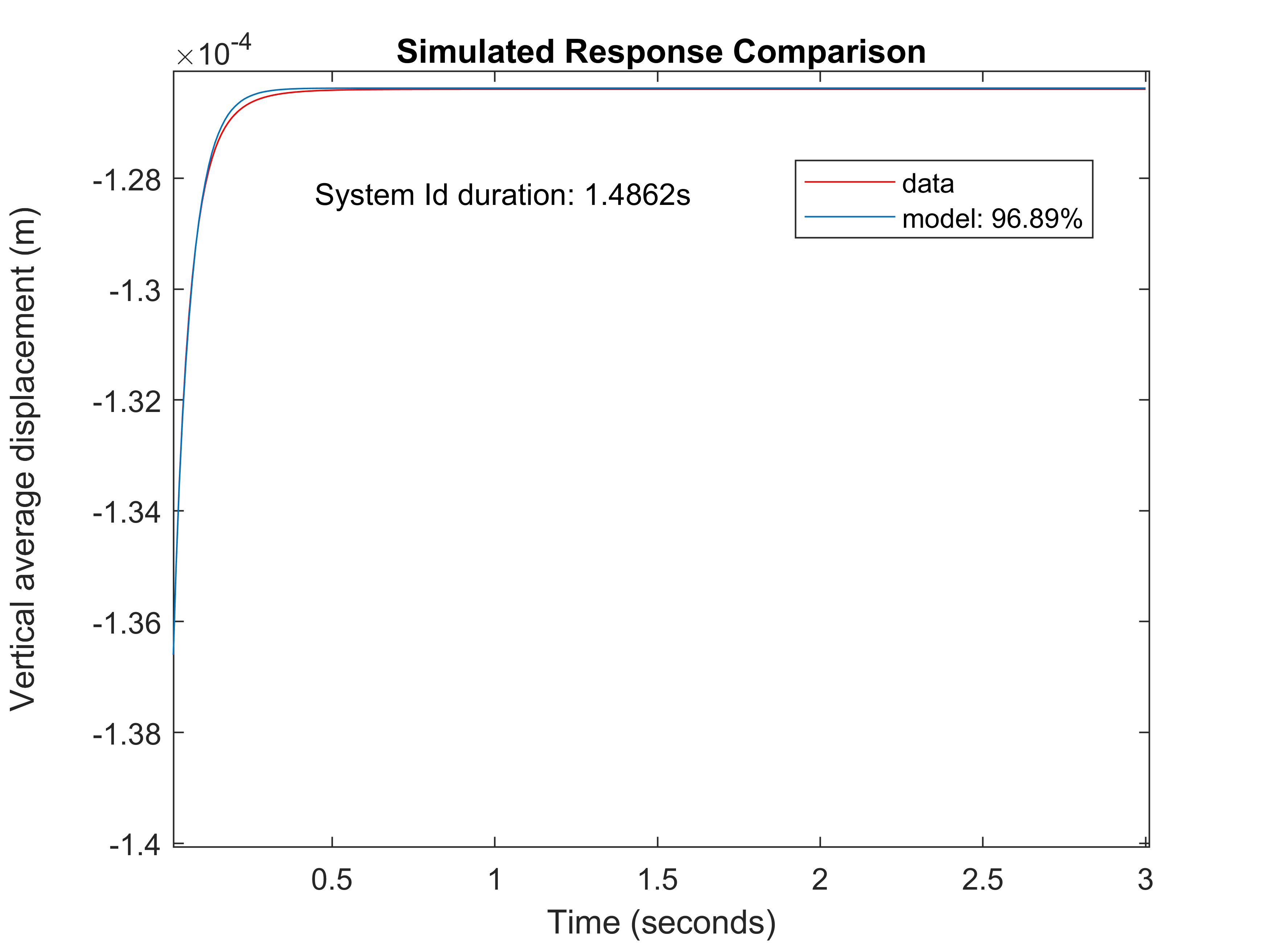}
	\caption{Comparison of simulation results between the reduced FEA model and the algebraic topological model of a  thermo-mechanical system.}
	\label{fig:ex-fit-coupled}
\end{figure}

To retrieve interpretability, we initialize a pair of connected oriented cell complexes representing a mechanical LPM (mass-spring-damper network) connected to a thermal LPM (resistance-conductance network) by a transformer (TF) 
(Fig. \ref{fig:ex-AT3}), where we label the $1-$cells with $L_3$, $L_6$ for masses, $L_2$, $L_5$ for springs, $L_1$, $L_4$ for dampers, $L_7$, $L_{10}$, $L_{11}$ for thermal resistors, and $L_8$, $L_9$ for thermal conductors. The Tonti diagram of generalized network systems \cite{wang2019topological}
is shown in Fig. \ref{fig:TD_generalized|paths} (a) and the paths traced to generate ODEs of the multi-domain lumped parameter systems are shown in Fig. \ref{fig:TD_generalized|paths} (b). We apply an ordinary least squares regression to obtain the optimal values of the lumped mass, stiffness coefficients, damping coefficients, thermal conductance, and thermal resistance. We compare the solutions of the HFM, the ROM and the optimal LPM in Fig. \ref{fig:ex-fit-coupled}, where the NRMSE of the parameter estimation is 3.11\% and the relative $\mathcal{H}_2-$error bound computed from (\ref{Eq：totalH2Bound}) against the HFM is 0.0124. Note that the mechanical vibration time scale in this case is much faster than the heat diffusion, so the displacement reaches steady-state value of $-1.393 \times 10^{ - 4}$ meters early on. The displacement visible in the figure is caused by thermal expansion.

\section{Conclusions}

We proposed a framework for automatically generating a family of surrogate models of physical systems with different cost-accuracy tradeoffs. We presented model-based, data-driven, and hybrid techniques that provide a priori guarantees of preserved properties including stability, convergence, and error bounds, as well as an ability to enforce an interpretable topological structure while assessing its impact on the error bound at a small computational cost. In particular, we used the SPARK+CURE algorithm to develop a first instance of a ROM for rapid simulation and data generation. We use this ROM to train another surrogate ROM (interpretable LPM) and measure the additional errors by solving algebraic Lyapunov equations. To systematically develop an interpretable LPM, we used Tonti diagrams of network theory to generate systems of ODEs from the presupposed topological structure and system identification to estimate the unknown constitutive parameters.  This approach avoids simulating the HFM and enables quantifying how well the assumed topology fits the ROM.

This framework can assist researchers and engineers in making decisions on the proper LOG to develop physical models of new engineering problems, and will open up new research directions. Possible areas of further research include extension to nonlinear systems using methods such as piecewise linearization, dynamic mode decomposition, and Koopman operator, and applying the ROM models to solve inverse problems (e.g., engineering design).

\section{Acknowledgments}
This material is based upon work supported by the Defense Advanced Research Projects Agency (DARPA) under Agreements No. HR00111990029 and HR00112090065.

\bibliographystyle{plainnat}    
\bibliography{bib}

\end{document}